# Modelling of the gravity compensators in robotic manufacturing cells

A. Klimchik*, Y. Wu*, S. Caro**, C. Dumas***, B. Furet*** and A. Pashkevich*

*Ecole des Mines de Nantes, 4 rue Alfred-Kastler, Nantes 44307 France (Tel: +33-251-85-83-00; e-mail: alexandr.klimchik@mines-nantes.fr, yier.wu@mines-nantes.fr, anatol.pashkevich@mines-nantes.fr).
**Centre National de la Recherche Scientifique, France (e-mail:stephane.caro@irccyn.ec-nantes.fr)
*** University of Nantes, France,  (e-mail: claire.dumas@univ-nantes.fr, benoit.furet@univ-nantes.fr)

**Abstract:** The paper deals with the modeling and identification of the gravity compensators used in heavy industrial robots. The main attention is paid to the geometrical parameters identification and calibration accuracy. To reduce impact of the measurement errors, the design of calibration experiments is used. The advantages of the developed technique are illustrated by experimental results .

*Keywords*: industrial robot, calibration, gravity compensator, geometrical modeling.

## 1. INTRODUCTION

Currently, aeronautic industry requires high-precision machining of huge aircraft component made from the high performance materials. For these applications, robots are quite attractive due to their large workspace that can be also easily extended. Another considerable advantage is their capability to process the parts with complex shape and geometry. However, processing of these materials causes essential cutting forces that are generated by tool-workpiece interaction during the material removal. Since these forces decrease the machining accuracy, the robot manufactures try to improve the robot stiffness by increasing the link cross-sections. This approach obviously leads to augmentation of robot link weights. So, the gravity forces applied to the manipulator components become non-negligible and also contribute to the position errors. To overcome this difficulty, the link weights are tended to be balanced by gravity compensators, which considerably complicate the stiffness modeling of these heavy manipulators.

The problem of stiffness modeling for the heavy manipulators with gravity compensators has been in the focus of rather limited number of works. In contrast, for conventional serial manipulators without gravity compensators, the problem has been studied by a number of authors that considered both industrial and medical robots with essential compliance in the links and joints (Meggiolaro 2005, Kövecses 2007). Relevant works are mainly based on the virtual joint method (VJM), which lumps the elastostatic properties of the robot components in virtual springs. To our knowledge, the stiffness modeling for the manipulators with gravity compensators has not been studied in detail yet. Currently, the main activity in this area focuses on the gravity compensator design (Takesue 2011, De Luca 2011). On the other hand, since the considered robots include closed loops induced by the compensators, some technique previously developed for the parallel manipulators can be adopted (Bouzgarrou 2004, Company 2005, Pashkevich 2011).

This paper focuses on the geometrical and stiffness modeling of the spring-based gravity compensators that can be integrated in a VJM-based stiffness model of a serial manipulator (Klimchik 2012). The main attention is paid to the identification of the model parameters and calibration experiment planning. The developed approach is confirmed by the experimental results that deal with the industrial robot employed in manufacturing of large-dimensional aircraft components.

To address these problems, the remainder of the paper is organized as follows. Section 2 presents the gravity compensator model. In Section 3, calibration methodology is presented. Section 4 proposes identification algorithms. Sections 5 is devoted to the experiment design. Experimental validation is presented in Section 6. Finally, Section 7 summarizes the main contributions.

## 2. GRAVITY COMPENSATOR MODEL

The mechanical structure of the gravity compensator under study is presented in Fig. 1.  The compensator incorporates a passive spring attached to the first and second links, which creates a closed loop that generates the torque applied to the second joint of the manipulator. Corresponding model is presented in Fig. 2, where the most essential geometrical parameters are denoted as $a_x$, $a_y$, $L$ and the compensator is described by the spring compliance $k_c$ and the preloading $s_0$. This design allows us to limit the stiffness model modification by incorporating in it the compensator torque and adjusting the virtual joint stiffness matrix that here depends on the second joint variable $q_2$ only.

The compensator geometrical model includes three node points $P_0$, $P_1$, $P_2$, where two distances $|P_1, P_2|$, $|P_0, P_2|$ are constants and the third one $|P_0, P_1|$ varies and depends on $q_2$. Let us denote them $L = |P_1, P_2|$, $a = |P_1, P_2|$, $s = |P_1, P_2|$. Besides, let us introduce the angles $\alpha$, $\varphi$ and the distances $a_x$ and $a_y$, whose geometrical meaning is clear from Fig. 2.

Using these notations, the variable $s$ describing the compensator spring deflection can be computed from the expression

$$s^2 = a^2 + L^2 + 2 \cdot a \cdot L \cdot \cos(\alpha - q_2) \quad (1)$$

which defines the function $s(q_2)$.

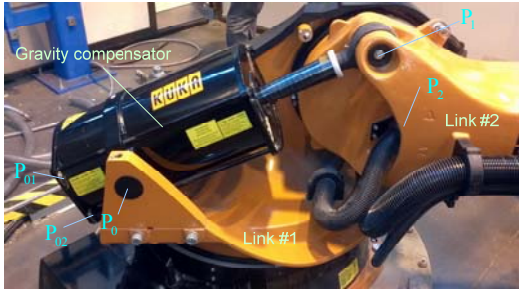

Fig. 1. Gravity compensator of robot KUKA KR-270 TM

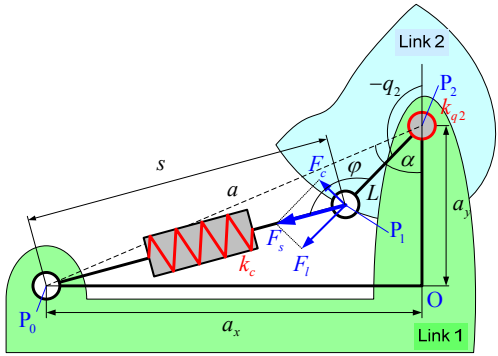

Fig. 2. Model of the gravity compensator

This mechanical design allows to balance the manipulator weight for any given configuration by adjusting the compensator spring preloading. It can be taken into account by introducing the zero-value of the compensator length $s_0$ corresponding to the unloaded spring. Under this assumption, the compensator force applied to the node $P_1$ can be expressed as follows

$$F_s = K_c \cdot (s - s_0) \quad (2)$$

where $k_c$ is the compensator spring compliance.

Further, the angle $\varphi$ between the compensator links $P_0P_1$ and $P_1P_2$ (see Fig. 2) can be found from the expression

$$\sin\varphi = a/s \cdot \sin(\alpha - q_2) \quad (3)$$

which allows us to compute the compensator torque $M_c$ applied to the second joint

$$M_c = K_c \cdot (1 - s_0/s) \cdot a \cdot L \cdot \sin(\alpha - q_2) \quad (4)$$

Upon differentiation of the latter expression with respect to $q_2$, the equivalent stiffness of the second joint (comprising both the manipulator and compensator stiffnesses) can be expressed as:

$$K_{\theta_2} = K_{\theta_2}^0 + K_c \cdot a\, L \cdot \eta_{q_2} \quad (5)$$

where the coefficient

$$\eta_{q_2} = \frac{s_0}{s}\left(\frac{a\,L}{s^2}\sin^2(\alpha - q_2) + \cos(\alpha - q_2)\right) - \cos(\alpha - q_2) \quad (6)$$

highly depends on the value of joint variable $q_2$ and the initial preloading in the compensator spring described by $s_0$.

Hence, using expression (5), it is possible to extend the classical stiffness model of the serial manipulator

$$\mathbf{K}_C = \left(\mathbf{J}_\theta^{(F)}(\mathbf{K}_\theta - \mathbf{H}_{\theta\theta})^{-1}\mathbf{J}_\theta^{(F)T}\right)^{-1} \quad (7)$$

by modifying the virtual spring parameters in accordance with the compensator properties. Here, $\mathbf{K}_C, \mathbf{K}_\theta$ are stiffness matrices in the Cartesian and joint spaces respectively, $\mathbf{J}_\theta^{(F)}, \mathbf{H}_{\theta\theta}$ are corresponding Jacobian and Hessian matrices (for more details see Klimchik 2012). While in the paper this approach has been used for the particular compensator type (spring-based, acting on the second joint), the similar idea can be evidently applied to other compensator types.

Summarizing this Section, it is worth mentioning that the geometrical and elastostatic models of a heavy manipulator with a gravity compensator should include some additional parameters ($a_x$, $a_y$, $L$ and $K_c$, $s_0$ for the presented case) that are usually not included in datasheets. For this reason, the following Sections focus on the identification of the extended set of manipulator parameters.

## 3. MODEL CALIBRATION METHODOLOGY

In contrast to the serial manipulator that can be treated as a principal mechanism of the considered robots, geometrical data concerning gravity compensators are usually not included in the technical documentation provided by the robot manufacturers. For this reason, this Section focuses on the calibration methodology of the geometrical parameters for the described above compensator mechanism (see Fig. 1).

The geometrical structure of the considered gravity compensator is presented in Fig. 2. Its principal geometrical parameters are denoted as $L$, $a_x$, $a_y$, where $a_x = a \cdot \cos\alpha$, $a_y = a \cdot \sin\alpha$ (see notation in Section III). As follows from the figure, the identification problem can be reduced to the determination of relative locations of points $P_0$ and $P_1$ with respect to $P_2$.

It is assumed that the measurement data are provided by the laser tracker whose "world" coordinate system is located at the intersection of the first and second actuated manipulator joints. The axes Y, Z of this system are aligned with the axes of joints #1 and #2 respectively, while the axis X is directed to ensure right-handed orthogonal basis. To obtain required data, there are several markers attached to the compensator mechanism (see Fig. 3). The first one is located at point $P_1$, which is easily accessible and perfectly visible (the center of the compensator axis $P_1$ is exactly ticked on the fixing element). In contrast, for the point $P_0$, it is not possible to locate the marker precisely. For this reason, several markers $P_{0i}$ are used that are shifted with respect to $P_0$, but located on the rigid component of the compensator mechanism (these markers are rotating around $P_0$ while the joint coordinate $q_2$ is actuated). It should be noted that for the adopted compensator geometrical model (which is in fact a planar one), the marker location relative to the plane XY is not significant, since the identification algorithm presented in the following sub-section will ignore Z-coordinate.

Using this setting, the identification problem is solved in two steps. The first step is devoted to the identification of the relative location of points $P_1$ and $P_2$. Here, for different values of the manipulator joint coordinates $\{q_{2i}, \ i=\overline{1,m}\}$, the laser tracker provides the set of the vectors $\{\mathbf{p}_1^i\}$ describing the points that are located in an arc of the circle. After matching these points with a circle, one can obtain the desired value of $L$ (circle radius) and the Cartesian coordinates $\mathbf{p}_2$ of the point $P_2$ (circle center) with respect to the laser tracker coordinate system.

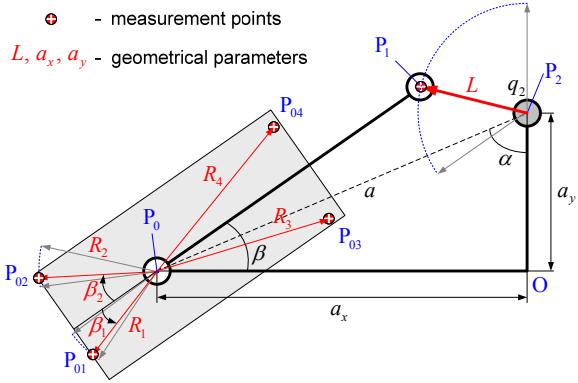

Fig. 3. Geometrical parameters of the gravity compensator and location of the measurement points labelled with markers

The second step deals with the identification of the relative location of points $P_0$ and $P_2$. Relevant information is extracted from two data sets $\{\mathbf{p}_{01}^i\}$ and $\{\mathbf{p}_{02}^i\}$ that are provided by the laser tracker while targeting at the markers $P_{01}$ and $P_{02}$. Here, the points are matched to two circle arcs with the same center (explicitly assuming that the compensator model is planar), which yields the Cartesian coordinates $\mathbf{p}_0$ of the point $P_0$, also with respect to the laser tracker coordinate system. Finally, the desired values $a_x$, $a_y$ are computed as a projection of the difference vector $\mathbf{a} = \mathbf{p}_2 - \mathbf{p}_0$ on the corresponding axis of the coordinate system.

As follows from the presented methodology, a key numerical problem in the presented approach is the matching of the experimental points with a circle arc. It looks like a classical problem, however, there is a particularity here caused by availability of additional data $\{q_{2i}\}$ describing relative locations of the points $\{\mathbf{p}_1^i\}$. This feature allows us to reformulate the identification problem and to achieve higher accuracy compared with the traditional approach.

## 4. IDENTIFICATION ALGORITHMS

The above presented methodology requires solution of two identification problems. The first one aims at approximating of a given set of points (with additional arc angle argument) with an arc circle, which provides the circle center and the circle radius. The second problem deals with an approximation of several sets of points by corresponding number of circle arcs with the centers on the same rotation axis. Let us consider them sequentially.

### 4.1 Algorithm #1

To match the given set of points $\{\mathbf{p}_i\}$ with additional set of angles $\{q_i\}$ with a circle arc, let us define the affine mapping

$$\mathbf{p}_i = \mu \mathbf{R} \mathbf{u}_i + \mathbf{t} \tag{8}$$

where $\mathbf{u}_i = [\cos q_i, \ \sin q_i, \ 0]^T$ denotes the set of reference points located on the unit circle whose distribution on the arc is similar to $\mathbf{p}_i$, $\mu$ is the scaling factor that defines the desired circle radius, $\mathbf{R}$ is the orthogonal rotation matrix, $\mathbf{t}$ is the vector of the translation that defines the circle center. It worth mentioning that such a formulation has an advantage (in the sense of accuracy) comparing to a traditional circle approximation and it is a generalization of Procrustes problem known from the matrix analysis.

Using equation (8), the identification can be reduced to the following optimization problem

$$F = \sum_{i=1}^{m} (\mathbf{p}_i - \mu \mathbf{R} \mathbf{u}_i - \mathbf{t})^T (\mathbf{p}_i - \mu \mathbf{R} \mathbf{u}_i - \mathbf{t}) \to \min_{\mu, \mathbf{R}, \mathbf{t}} \tag{9}$$

which should be solved subject to the orthogonality constraint $\mathbf{R}^T \mathbf{R} = \mathbf{I}$. After differentiation with respect to $\mathbf{t}$, the latter variable can be expressed as

$$\mathbf{t} = m^{-1} \sum_{i=1}^{m} \mathbf{p}_i - \mu \, m^{-1} \mathbf{R} \sum_{i=1}^{m} \mathbf{u}_i \tag{10}$$

That leads to the simplification of (9) to

$$F = \sum_{i=1}^{m} (\widehat{\mathbf{p}}_i - \mu \mathbf{R} \widehat{\mathbf{u}}_i)^T (\widehat{\mathbf{p}}_i - \mu \mathbf{R} \widehat{\mathbf{u}}_i) \to \min_{r, \mathbf{R}} \tag{11}$$

where

$$\widehat{\mathbf{p}}_i = \mathbf{p}_i - m^{-1} \sum_{i=1}^{m} \mathbf{p}_i; \qquad \widehat{\mathbf{u}}_i = \mathbf{u}_i - m^{-1} \sum_{i=1}^{m} \mathbf{u}_i \tag{12}$$

Further, differentiation with respect to $\mu$ yields to

$$\mu = \sum_{i=1}^{m} \widehat{\mathbf{p}}_i^T \mathbf{R} \widehat{\mathbf{u}}_i \Big/ \sum_{i=1}^{m} \widehat{\mathbf{u}}_i^T \widehat{\mathbf{u}}_i \tag{13}$$

So, finally, after relevant substitutions the objective function can be presented as

$$F = \sum_{i=1}^{m} \widehat{\mathbf{p}}_i^T \widehat{\mathbf{p}}_i - \left( \sum_{i=1}^{m} \widehat{\mathbf{u}}_i^T \widehat{\mathbf{u}}_i \right)^{-1} \left( \sum_{i=1}^{m} \widehat{\mathbf{p}}_i^T \mathbf{R} \widehat{\mathbf{u}}_i \right)^2 \to \min_{\mathbf{R}} \tag{14}$$

where the unknown matrix $\mathbf{R}$ must satisfy the orthogonality constraint $\mathbf{R}^T \mathbf{R} = \mathbf{I}$. Since the matrix $\mathbf{R}$ is included in the second term only, the problem can be further simplified to

$$F' = \sum_{i=1}^{m} \widehat{\mathbf{p}}_i^T \mathbf{R} \widehat{\mathbf{u}}_i = trace\left( \mathbf{R} \sum_{i=1}^{m} \widehat{\mathbf{u}}_i \widehat{\mathbf{p}}_i^T \right) \to \max_{\mathbf{R}} \tag{15}$$

and can be solved using SVD-decomposition of the matrix

$$\sum_{i=1}^{m} \hat{\mathbf{u}}_i \hat{\mathbf{p}}_i = \mathbf{U} \Sigma \mathbf{V}^T \quad (16)$$

where the matrices $\mathbf{U}, \mathbf{V}$ are orthogonal and $\Sigma$ is the diagonal matrix of the singular values. Further, using the same approach as for the Procrustes problem, it can be proved that the desired rotation matrix can be computed as

$$\mathbf{R} = \mathbf{V} \mathbf{U}^T \quad (17)$$

which sequentially allows to find the scaling factor $\mu$ defining the arc radius and the vector $\mathbf{t}$ defining the arc center.

*4.2 Algorithm #2*

The second problem aims at approximating of several point sets $\{\mathbf{p}_i^1\}, ..., \{\mathbf{p}_i^k\}$ by corresponding number of concentric circle arcs with the centers $\mathbf{p}_0^j$ on the rotation axis $\mathbf{n}$. It should be noted that the data set $\{q_i\}$ is not useful here, since the required angles $\{\beta_i\}$ are not measured directly and cannot be computed without having exact compensator geometry. In this case, the objective function can be written in the straightforward way

$$F = \sum_{j=1}^{k} \sum_{i=1}^{m} \left( R_j^2 - \left(\mathbf{p}_i^j - \mathbf{p}_0^j\right)^T \left(\mathbf{p}_i^j - \mathbf{p}_0^j\right) \right)^2 \to \min_{\mathbf{p}_0^j, R_j} \quad (18)$$

where $R_j$ denotes the $j$-th ark radius and $\mathbf{p}_0^j$ is the corresponding center point. However, for this formulation it can be easily proved that the optimization problem (18) does not lead to a unique solution. In fact, it gives the rotation axis passing through the center points $\mathbf{p}_0^j$, which can be expressed as

$$\mathbf{p}_0^j = \mathbf{p}_c + \xi_j \mathbf{n} \quad (19)$$

where $\mathbf{n}$ is the axis direction vector, $\mathbf{p}_c$ is a point belonging to the axis, and $\xi_j$ are corresponding scalar factors.

To solve the problem (18), first the objective function $F$ can be differentiate with respect to $R_j^2$ that yields the following expressions for the arc radii

$$R_j^2 = m^{-1} \sum_{i=1}^{m} \left(\mathbf{p}_i^j - \mathbf{p}_0^j\right)^T \left(\mathbf{p}_i^j - \mathbf{p}_0^j\right) \quad (20)$$

Further, after relevant substitution, the objective can be rewritten as

$$F = \sum_{j=1}^{k} \sum_{i=1}^{m} \left( 2\left(\mathbf{p}_c + \xi_j \mathbf{n}\right)^T \hat{\mathbf{p}}_i^j - \hat{s}_i^j \right)^2 \to \min \quad (21)$$

where

$$\hat{\mathbf{p}}_i^j = \mathbf{p}_i^j - m^{-1} \sum_{l=1}^{m} \mathbf{p}_l^j; \quad \hat{s}_i^j = \mathbf{p}_i^{jT} \mathbf{p}_i^j - m^{-1} \sum_{l=1}^{m} \mathbf{p}_l^{jT} \mathbf{p}_l^j \quad (22)$$

To solve the above mentioned ambiguity, additional objectives should be considered

$$R_j^2 \to \min_{R_j} \quad (23)$$

which leads to the following solution for the scalar parameter

$$\xi_j = \left(-\mathbf{p}_c + m^{-1} \sum_{i=1}^{m} \mathbf{p}_i^j\right)^T \mathbf{n} \quad (24)$$

Further, after differentiation (21) with respect to $\mathbf{p}_c$

$$\left(\sum_{j=1}^{k} \sum_{i=1}^{m} \hat{\mathbf{p}}_i^j \hat{\mathbf{p}}_i^{jT}\right) \mathbf{p}_c = \frac{1}{2} \sum_{j=1}^{k} \sum_{i=1}^{m} \hat{s}_i^j \hat{\mathbf{p}}_i^j \quad (25)$$

one can compute the point on the desired rotational axis as

$$\mathbf{p}_c = \frac{1}{2} \left(\sum_{j=1}^{k} \sum_{i=1}^{m} \hat{\mathbf{p}}_i^j \hat{\mathbf{p}}_i^{jT}\right)^{-1} \sum_{j=1}^{k} \sum_{i=1}^{m} \hat{s}_i^j \hat{\mathbf{p}}_i^j \quad (26)$$

The remaining unknown vector $\mathbf{n}$ can be obtained from the orthogonality constraints $\left(\mathbf{p}_i^j - \mathbf{p}_0^j\right)^T \mathbf{n} = 0$, $i = \overline{1, m}, j = \overline{1, k}$ that leads to the following optimization problem

$$f = \sum_{j=1}^{k} \sum_{i=1}^{m} \left( \left(\mathbf{p}_i^j - \mathbf{p}_0^j\right)^T \mathbf{n} \right)^2 \to \min_{\mathbf{n}} \quad (27)$$

that after substitution in it (19), (24) and (26) gives

$$f = \sum_{j=1}^{k} \sum_{i=1}^{m} \left(\hat{\mathbf{p}}_i^{jT} \mathbf{n}\right)^2 \to \min_{\mathbf{n}} \quad (28)$$

Further, differentiation (28) with respect to $\mathbf{n}$, the optimization problem reduces to the solving following homogeneous linear equation system

$$\sum_{j=1}^{k} \sum_{i=1}^{m} \hat{\mathbf{p}}_i^j \hat{\mathbf{p}}_i^{jT} \mathbf{n} = \mathbf{0} \quad (29)$$

Non-trivial solution of this system can be found using the singular value decomposition of the matrix $\sum_{j=1}^{k} \sum_{i=1}^{m} \hat{\mathbf{p}}_i^j \hat{\mathbf{p}}_i^{jT}$

$$\sum_{j=1}^{k} \sum_{i=1}^{m} \hat{\mathbf{p}}_i^j \hat{\mathbf{p}}_i^{jT} = \mathbf{U} \Sigma \mathbf{V}^T \quad (30)$$

where the vector $\mathbf{n}$ is the last column of the matrix $\mathbf{V}$.

It should be mentioned that practical application of the latter expressions is essentially simplified by the adopted assumption concerning orientation of the reference coordinate system, where the direction of the rotation axis $\mathbf{n}$ is close to Z-direction.

Hence, the developed algorithms allows us to identify the compensator geometrical parameters $L$, $a_x$, $a_y$ that are directly related to the above mentioned rotation center points $P_0$, $P_2$ and corresponding radii. Below they will be applied to the processing of the experimental data.

## 5 DESIGN OF CALIBRATION EXPERIMENTS

The main idea of the calibration experiment design is a set of robot configurations (as well as marker locations) that ensure the best identification accuracy. The key issue here is the ranging of different plans in accordance with the prescribed performance measure.

For the considered identification problem, the design variables are the set of angles $\{\mathbf{q}_{2i}\}$ and the marker locations. The objective functions to be minimized are computed via the covariance matrix that describes the identification errors for the geometrical parameters $L$ and $a$ to be estimated. Since two identification algorithms are independent, selection of the optimal configurations $\{\mathbf{q}_{2i}\}$ and marker locations can be considered sequentially.

Assuming that each experiment includes the additive measurement errors in the Cartesian coordinates $\varepsilon_i$, expression (13) allows us to present the variance of the parameter $\mu$ in the following way

$$\text{var}(\mu) = E\left(\sum_{i=1}^{m}\hat{\mathbf{u}}_i^T \mathbf{R}^T \varepsilon_i \sum_{i=1}^{m}\varepsilon_i^T \mathbf{R}\hat{\mathbf{u}}_i\right) \Big/ \left(\sum_{i=1}^{m}\hat{\mathbf{u}}_i^T\hat{\mathbf{u}}_i\right)^2 \quad (31)$$

where $E(.)$ denotes the expectation and the orthogonal matrix $\mathbf{R}$ defines the orientation of the base coordinate system. Following usual assumption concerning the measurement errors (independent identically distributed, with zero expectation and standard deviation $\sigma^2$ for each coordinate) that allows to present the covariance error matrix as $E(\varepsilon_i \varepsilon_i^T) = \sigma^2 \mathbf{I}$, the above expression (31) reduces to

$$\text{var}(\mu) = \sigma^2 \Big/ \left(\sum_{i=1}^{m}\hat{\mathbf{u}}_i^T\hat{\mathbf{u}}_i\right) \quad (32)$$

Further, it can be proved that $\sum_{i=1}^{m}\hat{\mathbf{u}}_i^T\hat{\mathbf{u}}_i = m$. So, the variance (32) does not depend on the angles $q_{2i}$. Thus, the identification accuracy for the parameter $L$ depends on the number of experiments only.

For the remaining geometrical parameter $a$, the identification error depends on the estimation precision of relative location of the points $P_2$ and $P_0$. Since relevant identification algorithms employ independent measurement data, the variance $\text{var}(a)$ can be computed as the sum of the traces of $\text{cov}(\mathbf{t}_2)$ and $\text{cov}(\mathbf{t}_0)$, where $\mathbf{t}_1$ and $\mathbf{t}_2$ are the vectors of Cartesian coordinates for the points $P_2$ and $P_0$.

For the point $P_2$, expression (10) leads to the following covariance matrix

$$\text{cov}(\mathbf{t}_2) = m^{-2} E\left(\sum_{i=1}^{m}\varepsilon_i \sum_{i=1}^{m}\varepsilon_i^T + \mu^2 \mathbf{R}\sum_{i=1}^{m}\mathbf{u}_i \sum_{i=1}^{m}\mathbf{u}_i^T \mathbf{R}^T\right) \quad (33)$$

which can be further simplified down to

$$\text{cov}(\mathbf{t}_2) = m^{-1}\sigma^2\left(\mathbf{I} + m^{-2}\sum_{i=1}^{m}\mathbf{u}_i \sum_{i=1}^{m}\mathbf{u}_i^T\right) \quad (34)$$

This simplification is based on the above derived expressions $E\left(\sum_{i=1}^{m}\varepsilon_i \sum_{i=1}^{m}\varepsilon_i^T\right) = m\sigma^2 \mathbf{I}$ and $E(\mu^2) = \sigma^2/m$ and on the assumption that z-axis of the coordinate system is directed along the second joint axis. Hence, for the point $P_2$, the optimization problem that is related to the design of calibration experiment, can be formulated as

$$F = \left(\sum_{i=1}^{m}\cos q_{2i}\right)^2 + \left(\sum_{i=1}^{m}\sin q_{2i}\right)^2 \to \min_{q_{2i}} \quad (35)$$

This problem should be solved taking into account joint limits of the industrial robot. In the case when the range of angles $q_2$ is over $\pi$, it is possible to achieve zero value of this objective since equations $\sum_{i=1}^{m}\cos q_{2i} = 0$ and $\sum_{i=1}^{m}\sin q_{2i} = 0$ are solvable. It should be noted similar equations arise in calibration experiment design for some robots without gravity compensators and have been studied in details in our previous work (Klimchik 2011).

For the point $P_0$, similar expression includes a set of the angles $\beta_i$ that can be recomputed to the joint angles $q_{2i}$ requires for the manipulator control (see Fig. 3). Here, it is reasonable to find optimal marker locations on the rigid part of the gravity compensator. It can be proven that using these assumptions, the design of experiment reduces to the following optimization problem

$$F = \left(\sum_{j=1}^{k}\cos \beta_j\right)^2 + \left(\sum_{j=1}^{k}\sin \beta_j\right)^2 \to \min_{\beta_j} \quad (36)$$

where $\beta_j$ are the angles around the point $P_0$ between the compensator spring and $j$-th marker location. It is clear that in this case the best solution is produced by similar equations $\sum_{j=1}^{k}\cos \beta_j = 0$ and $\sum_{j=1}^{k}\sin \beta_j = 0$, but contrary to (35), this problem can be easily solved by locating the markers on the opposite sides of the compensator rotation axis.

Thus, the calibration experiment design that produces the sets of the optimal manipulator configurations and the marker locations described by the variables $\{q_{2i}\}$ and $\{\beta_j\}$ respectively, is reduced to the solution of the above presented trigonometric equations that allows us essentially increase the calibration accuracy.

## 6 EXPERIMENTAL RESULTS

To demonstrate efficiency of the developed technique, the experimental study has been carried out. The experimental setup employed the robot KR-270 and the Leica laser tracker, which allowed us to measure the Cartesian coordinates of the markers attached to the compensator elements with the accuracy of 10 μm (see Figs 1, 3). Six different manipulator configurations where considered that differed in the value of the joint angle $q_2$ and three markers has been used. The experimental data are presented in Table 1.

These data has been processed using the developed identification algorithm presenting in the Section 4. The obtained values for the parameters of interest $L$, $a_x$, $a_y$ are given in Table 2, which also includes the identification errors computed using the Gibbs sampling technique.

In addition, there were evaluated the elastostatic properties of the gravity compensator. Corresponding curves describing influence of the compensator on the equivalent stiffness of the manipulator joint (see eq. (6)), are presented in Fig. 4. They demonstrate essential non-linearity of the compensator impact throughout of the robot workspace, which, in addition, highly depend on the spring preloading $s_0$.

**Table 1. EXPERIMENTAL DATA**

| $q_2$ [deg] | $P_1$ | | $P_{01}$ | | $P_{02}$ | |
|---|---|---|---|---|---|---|
| | x, [mm] | y, [mm] | x, [mm] | y, [mm] | x, [mm] | y, [mm] |
| -0.01 | -31.84 | 183.86 | -872.10 | -125.38 | -813.50 | -255.59 |
| -30 | -118.44 | 143.42 | -872.30 | -126.07 | -813.33 | -256.18 |
| -60 | -173.30 | 65.12 | -872.50 | -109.90 | -825.09 | -244.64 |
| -90 | -181.76 | -30.14 | -868.43 | -78.20 | -844.66 | -219.04 |
| -120 | -141.45 | -116.82 | -858.90 | -47.60 | -859.43 | -190.44 |
| -145 | -78.10 | -165.47 | -852.53 | -33.68 | -864.66 | -176.01 |

**Table 2. GEOMETRICAL PARAMETERS**

| | L, [mm] | $a_x$, [mm] | $a_y$, [mm] |
|---|---|---|---|
| value | 184.72 | 685.93 | 120.30 |
| accuracy | ±0.06 | ±0.70 | ±0.69 |

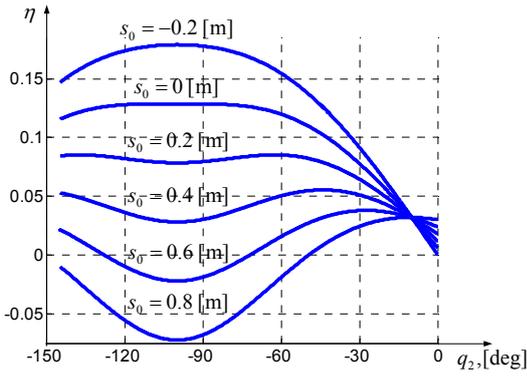

Fig. 4. Variation of the gravity compensator impact on the equivalent stiffness of the second joint

Another set of experiments have been carried out to identify the elastostatic properties of the considered manipulator. There were considered 15 different configurations, which have been found taking into account physical constraints that are related to the joint limits, work-cell obstacles and safety reasons. To ensure identification accuracy for each configuration, the experiments were repeated three times. In total, 405 equations were considered for the identification, from which 7 physical parameters have been obtained. Corresponding values of the elastostatic parameters for the gravity compensator and for the manipulator are presented in Table 3, where $k_i$ denotes the $i$-th joint compliance. There were also computed the identification errors (using the Gibbs sampling technique). As follows from Fig. 5, the gravity compensator essentially reduces the equivalent compliance of the second joint (compared to the serial manipulator without the gravity compensator).

**Table 3. ELASTOSTATIC PARAMETERS**

| Parameter | value | accuracy |
|---|---|---|
| $k_c$, [rad×μm/N] | 0.144 | ±0.031 (21.5%) |
| $s_0$, [mm] | 458 | ±27 (5.9%) |
| $k_2$, [rad×μm/N] | 0.302 | ±0.004 (1.3%) |
| $k_3$, [rad×μm/N] | 0.406 | ±0.008 (2.0%) |
| $k_4$, [rad×μm/N] | 3.002 | ±0.115 (3.8%) |
| $k_5$, [rad×μm/N] | 3.303 | ±0.162 (4.9%) |
| $k_6$, [rad×μm/N] | 2.365 | ±0.095 (4.0%) |

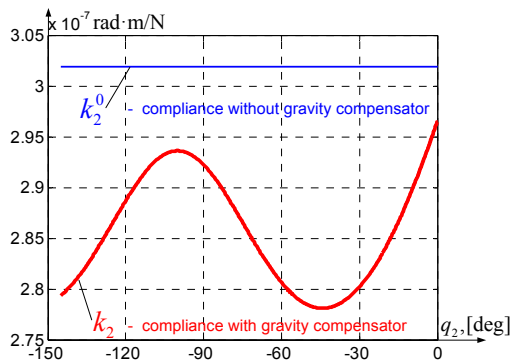

Fig. 5. Compliance of equivalent non-linear spring in the second joint

## 7. CONCLUSIONS

The paper presents a new approach for modeling and calibration of heavy industrial robots with gravity compensators. It proposes a methodology and data processing algorithms for the identification of the gravity compensator geometrical parameters. To increase the identification accuracy, the design of experiments has been used aimed at proper selection of the measurement configurations and marker point locations. The advantages of the developed techniques are illustrated by experimental study of the industrial robot Kuka KR-270, for which the model parameters of the gravity compensator have been identified.

ACKNOWLEDGMENT

The work presented in this paper was partially funded by the ANR, France (Project ANR-2010-SEGI-003-02-COROUSSO). The authors also thank Fabien Truchet, Guillaume Gallot, Joachim Marais and Sébastien Garnier for their great help with the experiments.